\documentclass[letterpaper, 10 pt, conference]{ieeeconf}  
\usepackage[utf8]{inputenc}

\IEEEoverridecommandlockouts

\usepackage{cite}
\usepackage{amsmath,amssymb,amsfonts}
\usepackage{algorithmic}
\usepackage{graphicx}
\usepackage{textcomp}
\usepackage{array}

\usepackage{graphicx}
\usepackage{float} 
\usepackage{blindtext}
\usepackage{afterpage}
\usepackage{hhline,caption}
\usepackage{hyperref}
\usepackage{comment}

\usepackage[utf8]{inputenc}
\usepackage{subcaption}
\usepackage{xcolor}

\usepackage[ruled]{algorithm2e}

\author{Wen Fan*, Haoran Li*, Dandan Zhang*
}
\begin{document}

\title{\LARGE \bf MagicTac: A Novel High-Resolution 3D Multi-layer Grid-Based \\Tactile Sensor
\thanks{W. Fan and D. Zhang are with the Department of Bioengineering, Imperial College London. H. Li is with the School of Engineering Mathematics and Technology, and Bristol Robotics Laboratory, University of Bristol. Corresponding: Dandan Zhang, d.zhang17@imperial.ac.uk}}

\maketitle

\begin{abstract}
Accurate robotic control over interactions with the environment is fundamentally grounded in understanding tactile contacts. In this paper, we introduce MagicTac, a novel high-resolution grid-based tactile sensor. This sensor employs a 3D multi-layer grid-based design, inspired by the Magic Cube structure. This structure can help increase the spatial resolution of MagicTac to perceive external interaction contacts. Moreover, the sensor is produced using the multi-material additive manufacturing technique, which simplifies the manufacturing process while ensuring repeatability of production. Compared to traditional vision-based tactile sensors, it offers the advantages of i) high spatial resolution, ii) significant affordability, and iii) fabrication-friendly construction that requires minimal assembly skills. We evaluated the proposed MagicTac in the tactile reconstruction task using the deformation field and optical flow. Results indicated that MagicTac could capture fine textures and is sensitive to dynamic contact information. Through the grid-based multi-material additive manufacturing technique, the affordability and productivity of MagicTac can be enhanced with a minimum manufacturing cost of £4.76 and a minimum manufacturing time of 24.6 minutes.

\end{abstract}

\IEEEpeerreviewmaketitle

\section{Introduction}

Recently, Vision-Based Tactile Sensors (VBTSs) have attracted considerable attention for their ability to capture high-resolution tactile information without substantially increasing equipment or maintenance costs.
Among current VBTSs, the Marker Displacement (MD)-based method has been widely used to map tactile information \cite{li2023marker}. This design leverages the motions of marker patterns, either positioned on the elastomer's surface or embedded within the sensor's skin, to reflect contact deformation. Such features can be detected by cameras and subsequently analyzed by computer vision-based algorithms. Serving as an essential component of MD-based VBTSs, the markers can be recognized as signal transducing patterns\cite{scharff2022rapid}. For example, the TacTip sensor \cite{lepora2021soft} uses a set of 3D-printed, biomimetic internal pins, which are known as pin-shaped markers. These markers are designed to magnify and reflect contact information when interacting with the environment. However, its capability to reconstruct fine texture features during contact is limited due to the relatively low density of markers. 

\begin{figure}[t]
    \centering
    \captionsetup{font=footnotesize,labelsep=period}
    \includegraphics[width = 0.9\hsize]{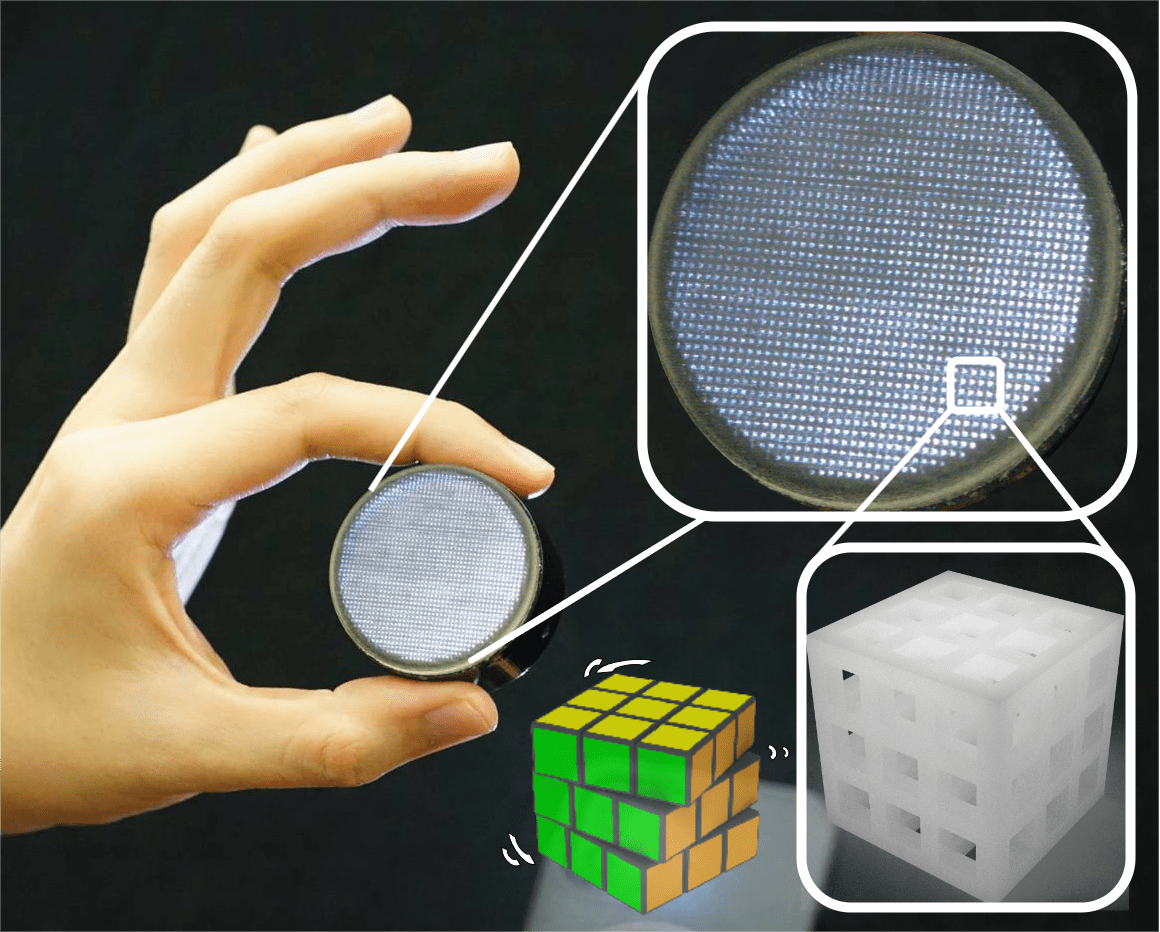}
    \caption{Overview of MagicTac: this design inspired by the magic cube, features a 3D multi-layer, grid-based tactile sensor. The 3D grid-based structure is visible through the mesh-like texture of the skin surface.}
    \label{first glance}
    \vspace{-0.6cm}
\end{figure}

In contrast, GelSight-type sensors \cite{yuan2017gelsight}, such as GelTip \cite{gomes2020geltip}, GelSlim \cite{taylor2022gelslim}, utilize reflective surfaces to capture fine-texture sensing. These sensors show better feature reconstruction capabilities compared to MD-based VBTSs. Furthermore, several GelSight-type sensors introduce ink dots as markers to enhance shear contact sensitivity\cite{yuan2017gelsight,sato2009finger}. However, this method of embedding 2D planar markers into the elastomer necessitates several extra manual steps in the fabrication process.
In addition, such fabrication is not only time-consuming and labor-intensive but also may introduce inconsistencies in the markers' size, shape, and arrangement.



Most of the existing VBTSs are facing challenges: For MD-based VBTSs, their marker patterns, designed as conventional dot or pin-shaped arrays, have seen limited improvement in the efficiency of tactile information mapping. Additionally, there will be a manufacturing constraint to increase the density of these sparse markers, affecting their enhancement of spatial resolution and contact sensitivity. Reflective coating-based VBTSs offer fine texture resolution and dynamic contact feature capture. But their complex structure complicates manufacturing, making it hard to produce inexpensive and stable products. Other recently developed VBTSs, such as DTac\cite{lin2023dtact}, have shown performance improvements over previous VBTSs. However, these novel sensors also may struggle with stable, large-scale manufacturing.

Here, we propose a novel tactile sensor with a 3D multi-layer grid-based structure, called 'MagicTac', as shown in Fig. \ref{first glance}. The innovative grid-based design offers MagicTac several unique capabilities for tactile sensing performance. Compared to MD-based VBTSs, grid cells are uniformly distributed in multi-layers throughout the MagicTac, not just on a surface, helping to improve spatial resolution. Unlike VBTSs that rely on both coating layers and markers, MagicTac can capture fine texture information and dynamic information without requiring additional structures. Furthermore, MagicTac stands out from other novel VBTSs with simplified manufacturing processes. Its enhanced affordability and repeatability are achieved by reducing post-processing and assembly costs through multi-material additive manufacturing. The \textbf{main contributions} of this paper are listed below:

\begin{itemize}
\item We propose the 3D multi-layer grid-based structure for tactile sensor construction, which aims to further enhance the tactile sensing of VBTS.
\item We utilize the multi-material additive manufacturing technique to simplify the fabrication process, which leads to reduced fabrication time and cost.
\end{itemize}


\section{Related Work}
\subsection{Design of VBTS}
\label{marker}
GelSight-type sensors, including GelTip \cite{gomes2020geltip}, DIGIT\cite{lambeta2020digit}, OmniTact\cite{padmanabha2020omnitact}, and others\cite{romero2020soft}, utilize reflective layers\cite{yuan2017gelsight} for accurate tactile reconstruction. Additionally, the integration of markers\cite{taylor2022gelslim, do2023densetact, zhang2023gelstereo} further enhances their dynamic perception capabilities. Some other popular VBTSs capture contact information primarily through marker displacement \cite{li2023marker}. The majority employ single-layer marker patterns, either embedded within elastomer \cite{yang2021enhanced} or printed directly onto the sensor's surface \cite{lepora2022digitac}. The GelForce \cite{sato2009finger} innovates by embedding markers in two layers for traction field measurement. Recent advancements also include double-layer MD-based tactile sensing, which leverages subtractive color mixing for tactile perception\cite{lin2019sensing,lin2020curvature}. Additionally, FingerVision\cite{yamaguchi2016combining,yamaguchi2021fingervision} integrates markers within a clear elastomer, which incorporates visual features into tactile perception. 


\subsection{Fabrication of VBTS}

According to \cite{abad2020visuotactile,zhang2022hardware}, the lens and housing base of VBTSs are typically created using laser cutting and 3D printing techniques. However, their used elastomer is often produced manually.  Silicone, favored for its excellent elasticity and transparency, is the most commonly used material. It is made by mixing A/B solutions and then casting in moulds\cite{lambeta2020digit,padmanabha2020omnitact,yamaguchi2016combining,yamaguchi2021fingervision}. The process of mould design and silicone preparation requires intricate procedures and manual skills\cite{lin2023dtact,lin20239dtact}, which constrains the design and fabrication flexibility and efficiency. For GelSight-type sensors, the reflective coating is typically applied through airbrushing\cite{lambeta2020digit, taylor2022gelslim, wang2021gelsight} or sputtering \cite{jiang20183}, which is a tedious process.

For MD-based VBTSs, typical approaches to embedding single-layer markers include silkscreen printmaking \cite{sato2009finger}, direct casting \cite{winstone2013tactip}, adhered markers to mold \cite{sakuma2018universal}, and using multi-material additive manufacturing \cite{ward2017exploiting}. Producing double-layer MD-based sensors involves a more complex process, requiring additional steps for assembly \cite{scharff2022rapid,lin2020curvature}.
TacTip-type sensors\cite{lepora2021soft,lepora2022digitac} benefit from 3D printing technology, allowing markers to be fabricated alongside the sensor skin. It significantly enhances fabrication accuracy and efficiency over traditional methods. However, this approach still necessitates the manual task of filling the internal cavity with soft silicone, adding to the production effort.

To simplify the fabrication process of VBTSs, we propose using the grid-based multi-material additive manufacturing technique (denoted as integral printing below) to replace the manual fabrication process, without any silicone casting, marker building, or skillful assembly. This ensures high repeatability and scalability of VBTS products by eliminating fabrication and assembly errors caused by humans.

\begin{figure}[!htbp]
    \centering
    \includegraphics[width = 0.65\hsize]{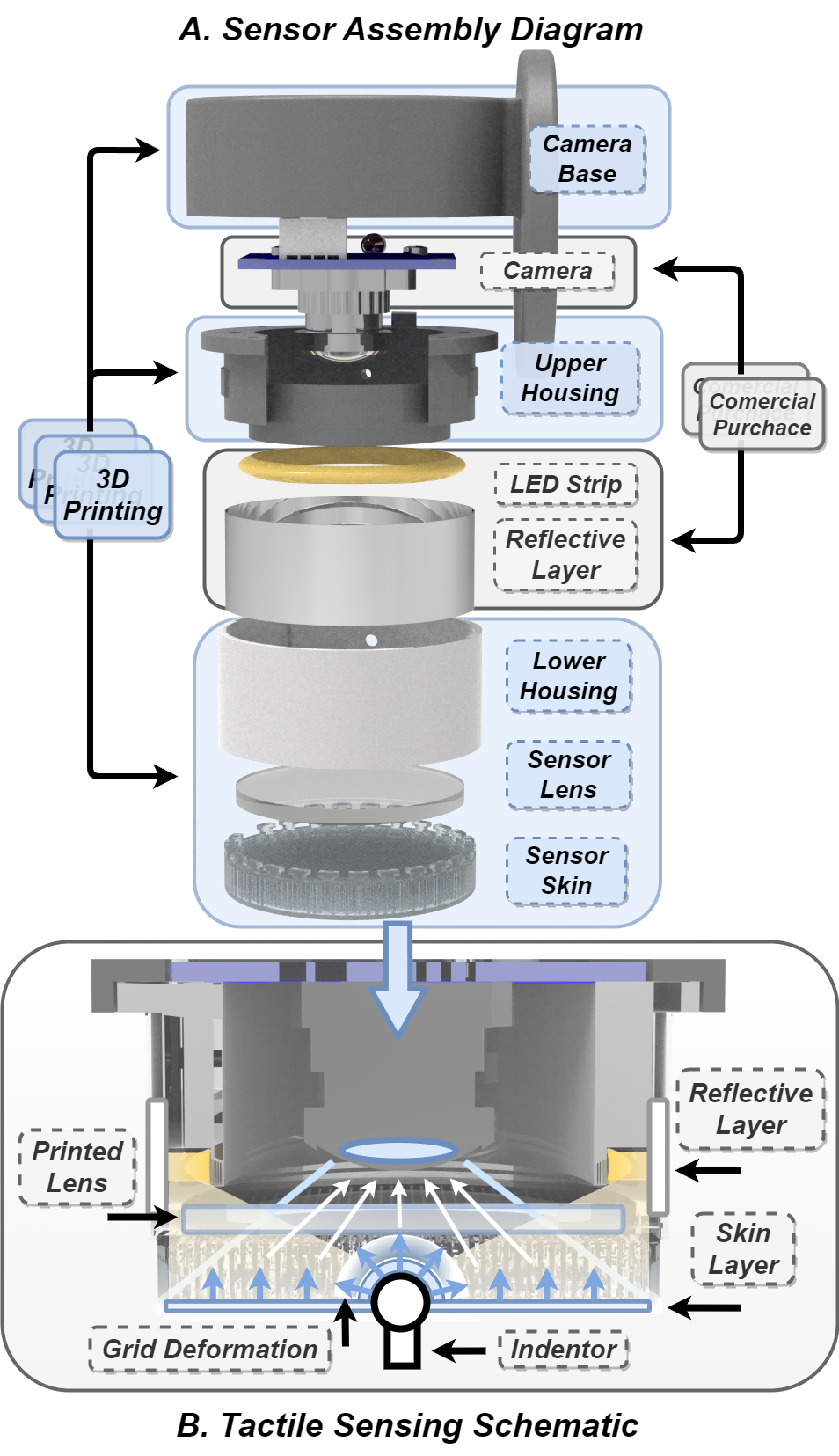}
    \captionsetup{font=footnotesize,labelsep=period}
    \caption{A: MagicTac assembly diagram, an exploded view of the subassemblies. B: Tactile sensing schematic, with a section view of MagicTac.}
    \label{explod}
    \vspace{-0.3cm}
\end{figure}

\begin{figure*}[!htbp]
    \centering
    \captionsetup{font=footnotesize,labelsep=period}
    \includegraphics[width = 0.9\hsize]{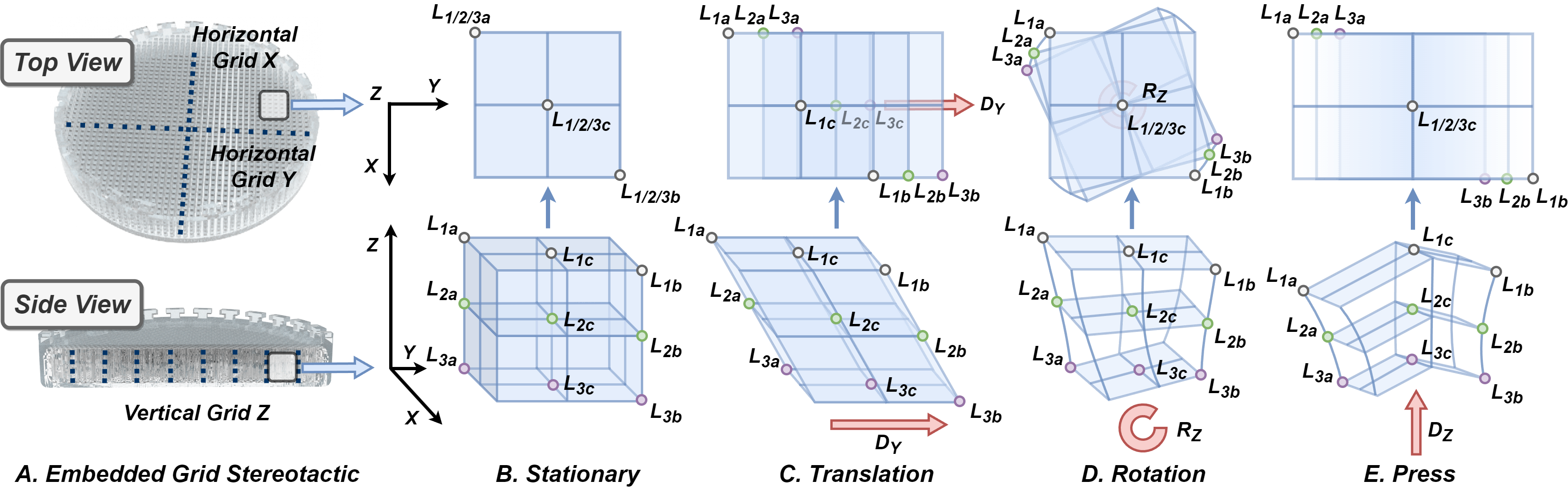}
    \caption{A: The schematic of 3D multi-layer grid-based structure, shown in top and side view. B-E: These examples illustrate the deformation in multiple contact situations, including stationary, translation, rotation, and press. The ability of this structure to map tactile features derives from the relative motion between the different layered grid cells, and their inherent elasticity. This characteristic forms the basis for the analogy to the \textbf{Magic Cube}.}
    \label{3d_marker}
    \vspace{-0.5cm}
\end{figure*}

\section{Design and Fabrication}

\subsection{Structure and Fabrication}
MagicTac consists of two modules, including a perception module, and a contact module. The assembled MagicTac is around 41$mm$ high with a maximum diameter (camera-mounted base) of 49.2$mm$.

\subsubsection{Perception Module}

The perception module is comprised of a camera, an LED ring, and a reflective layer, as shown in Fig.~\ref{explod} (A). We employ a USB camera characterized by a wide field of view (FOV) spanning 160 degrees. This camera can capture images at a resolution of 800x600 pixels and operate at a high frame rate of 60 frames per second (FPS). To maintain consistent and reliable illumination, we incorporate a readily available commercial LED ring and an extra reflective layer, made of common tinfoil. 

As shown in Fig.~\ref{explod}(B), the reflective layer aids the projecting light from the ring-shaped LED strip onto the sensor surface, while simultaneously minimizing significant reflections on the lens. This uniformly emitted light refracts upon entering the lens. A segment of this refracted light has an angle of incidence that exceeds the critical angle, enabling it to penetrate the lens medium and reach the 3D multi-layer grid-based structure inside. After this interaction, the light is further scattered, refracted, and reflected by the embedded grid and is then captured by the camera for imaging. This process creates a uniformly illuminated environment within the entire cavity of MagicTac, ensuring optimal conditions for tactile image acquisition.

\subsubsection{Contact Module}
The contact module consists of three critical components: the skin, which encompasses an out layer and the embedded 3D multi-layer grid-based structure, a transparent lens, and a housing part. These components were unifiedly manufactured using the Stratasys J826 printer, which is capable of printing a diverse range of materials varying from flexible to rigid and from opaque to transparent. Three types of materials are used for MagicTac, including the Agilus30 (a rubber-like flexible material), Vero (a multi-color rigid material), and support material (ultra-soft).

As shown in Fig.~\ref{explod}, the skin layer and embedded grid are integrated into a single entity, constructed using Agilus30 Clear (transparent). 
This particular embedded grid is composed of support material with an internal mesh made of Agilus30 Clear. The underlying principle employs the more rigid Agilus30 to form the grid's skeleton, which ensures strength and durability. This approach ensures the composite structure maintains sufficient flexibility for deformation without compromising the integrity of the grid structure. 

The thickness of the skin layer is adjustable from 0.3$mm$. For MagicTac developed in this study, the thickness was determined to be 0.6$mm$, a specification designed to strike an optimal balance between softness and robustness.
Subsequent average measurements obtained from several samples, using a durometer, demonstrated that this adjustment significantly lowered the MagicTac's stiffness from 30A to 17A. This reduction in stiffness illustrates the tangible impact of the thickness adjustment on the material's mechanical properties.


As shown in Fig. \ref{3d_marker}(A), the sensor skin possesses a diameter of 39.5$mm$ with a 6$mm$ height. Along the direction of skin diameter, a total of 42 grid cells are evenly distributed, and around 1385 grid cells for each layer. The horizontal dimension of the single grid cell is approximately 0.6$mm$, with a 0.3$mm$ separation existing between adjacent cells. It is similar in the vertical direction so each cell can be seen as a cube. It is calculated that there are about 10,000 grid cells in the entire skin. In contrast, the transparent lens with a diameter of 38.5$mm$ and thickness of 2.5$mm$ is fabricated by VeroClear (transparent). The housing part with a height of 19.5$mm$, outer diameter of 41$mm$, and wall thickness of 2.5$mm$ is made from VeroBlack (opaque).

The deformation principle of the 3D multi-layer grid-based structure is analyzed in Fig.~\ref{3d_marker} (B-E). As an example, a 2×2x2 grid structure with eight flexible cells, which contains the upper, middle, and lower plane layers $L_{1}$, $L_{2}$, $L_{3}$. Each layer has been defined with three virtual corner points, two diagonal vertexes $a$, $b$, and one center point $c$. In the stationary case, looking down from the top view, the corner points on each layer overlap each other. Once a shear translation $D_{Y}$ along the Y-axis is applied, points that were originally obscured will thus become visible again and move along the translation direction. Twisting $R_{Z}$ along the Z-axis leads to similar results, with the difference that the center points should still overlap. 
Applying an upward press ($D_{Z}$) along the Z-axis significantly increases the deformation complexity of the grid-based structure. It induces an upward bulge at the interface between adjacent columns of cells and causes horizontal stretching in the upper layer's cells due to their elasticity. 
This is crucial to understanding the adaptability and resilience of the 3D multi-layer grid-based structure under various mechanical stresses.

\begin{figure*}[t!]
		\centering
  \captionsetup{font=footnotesize,labelsep=period}
		\begin{tabular}[b]{c}
\hspace{0cm}\includegraphics[width=0.92\textwidth,trim={0 0 0 0},clip]{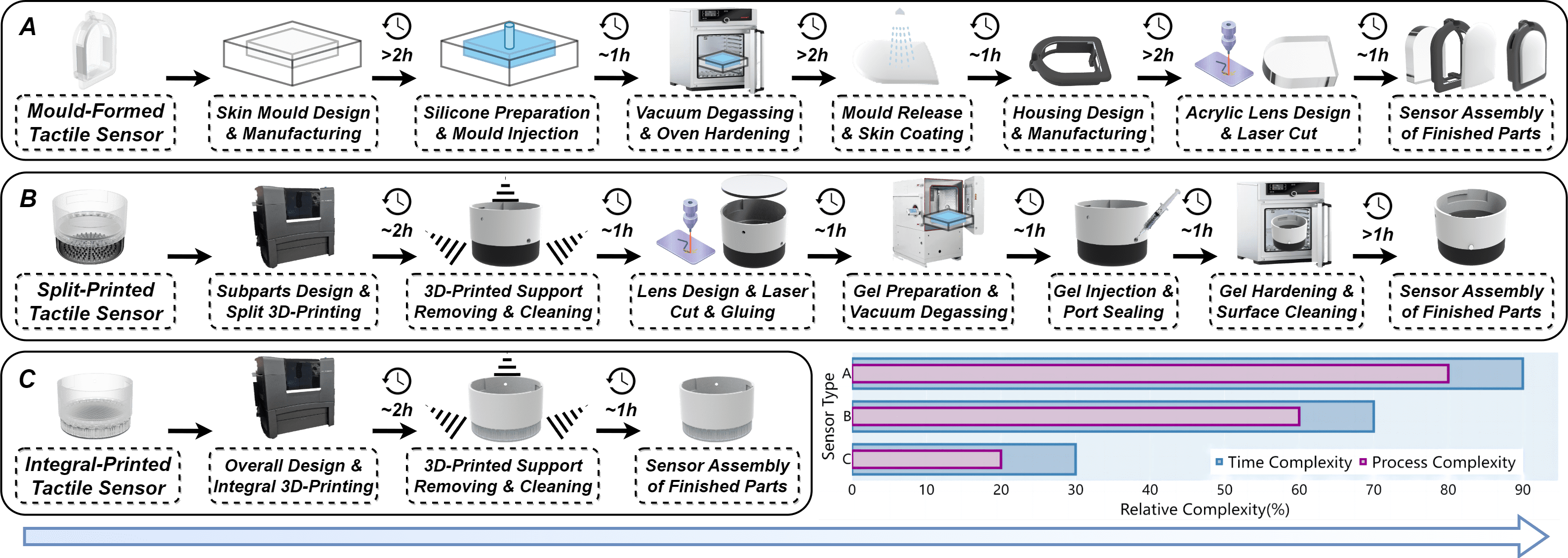} \\
\end{tabular}
\caption{Comparison of VBTS fabrication processes using different manufacturing methods. A: Mould-formed method has the highest time cost and process complexity; B: Split-printing way eliminates the cost of mould making, but lens assembly and internal gel injection still make the overall process cumbersome; C: Integral-printing method requires tiny pre-processing and post-processing work, thereby reducing time and labor costs significantly.}
		\label{fabrication}%
		\vspace{-1.5em}
\end{figure*}

\subsubsection{Fabrication Process}

MagicTac fabrication details are illustrated in Fig. \ref{fabrication}, with the comparison to the mould-formed tactile sensors \cite{yuan2017gelsight, lin2020curvature, lambeta2020digit, yamaguchi2016combining} and the split-printed tactile sensors \cite{scharff2022rapid, lepora2021soft}. As defined in \cite{zhang2023novel}, the sensor cost should include both hardware cost and manufacturing cost. However, typically, the manufacturing cost could exceed that of the hardware. Additionally, a general and reproducible process flow is important for VBTS manufacturing. The mould-formed method entails the highest complexity, as it requires creating a separate mould for the elastomer, which does not constitute the sensor's body itself. If different sizes, shapes, or quantities of sensors are desired, extra moulds will need to be made in advance to meet such requirements. This limitation significantly reduces flexibility, referred to as `mould dependency'. Moreover, another disadvantage of this method is the challenge of consistently producing markers with uniform shapes and sizes, even for simple 2D arrays. 

In contrast, the split-printing technique addresses the aforementioned challenges by manufacturing the elastic skin, markers, and housing components separately, which significantly lowers production costs. However, it introduces difficulty in removing printing supports from around the delicate markers embedded within the skin
Additionally, further procedures such as lens cutting and assembly, alongside gel preparation and injection, impose substantial time demands on researchers. As compared, the proposed integral printing approach offers a simpler alternative, negating the need for these additional fabrication stages.
The primary requirement with this method is the removal of a minimal amount of soft support material at the base of the component, after which the sensors are immediately ready for use. 


In summary, the integral-printing method presents several significant advantages:
i) Cost Efficiency: It reduces production costs by limiting the necessity for comprehensive post-processing and assembly; ii) Scalability: It facilitates the mass production of sensors, efficiently producing large volumes without quality sacrifice; iii) Consistency: It guarantees uniform quality and low variability among products, essential for applications demanding precise sensor functionality;
iv) Versatility: It offers customization options for sensors in various layouts, shapes, and sizes, catering to specific research or application requirements.
Additionally, the availability of rapid and superior 3D printing services from online providers increases accessibility, allowing all researchers to design and customize their own sensors precisely, even in the absence of personal printing resources.

\begin{figure}[!htbp]
    \centering
    \includegraphics[width = 0.8\hsize]{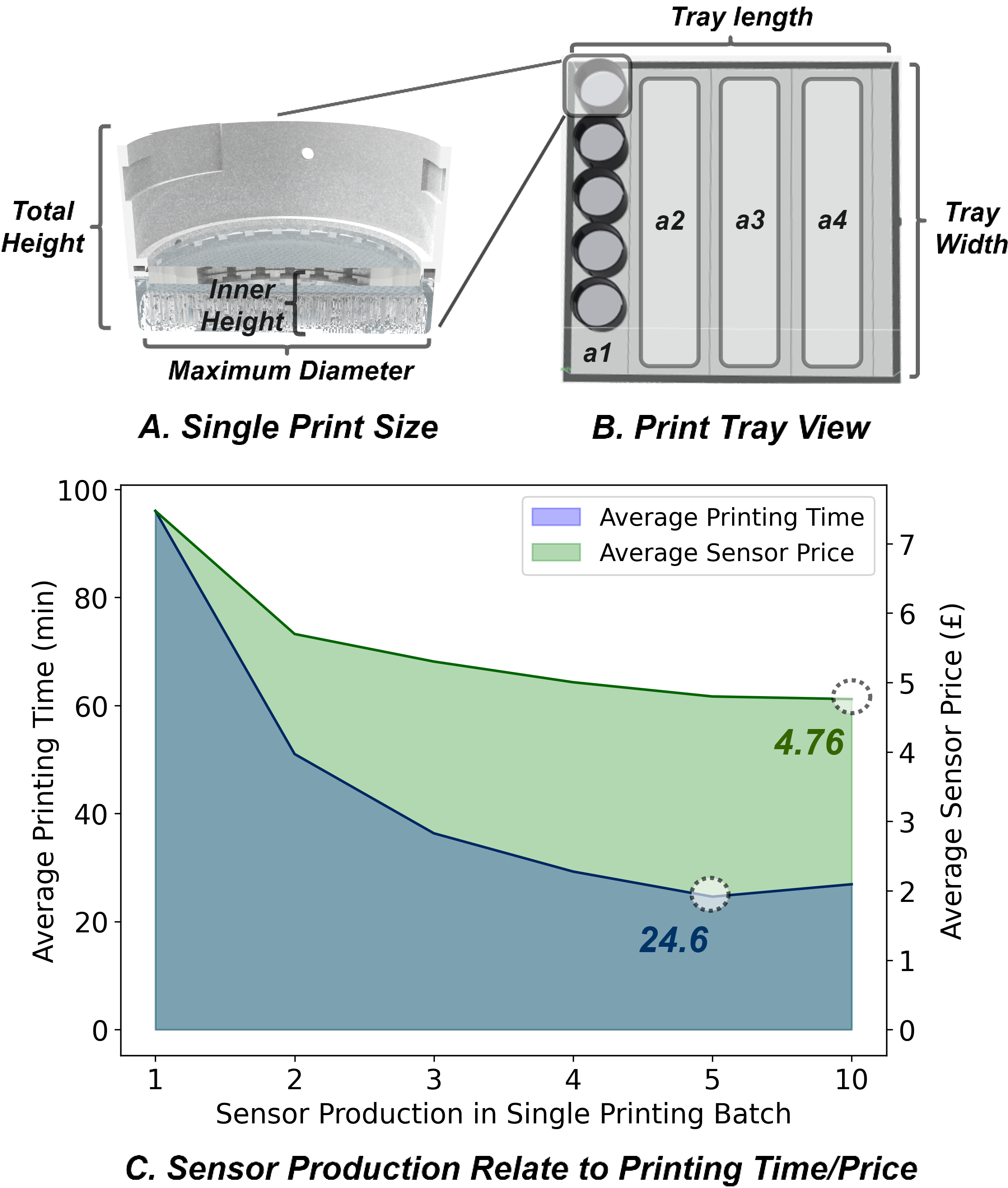}
\captionsetup{font=footnotesize,labelsep=period}
    \caption{A: The printing size of single MagicTac. B: Printing batch of MagicTacs on the print tray. C: The trends of average printing time and price related to batch production.}
    \label{print_time_scene}
    \vspace{-1.5em}
\end{figure}

\subsection{Affordability and Productivity Evaluation}
Affordability and productivity significantly influence the manufacturing of VBTS. Leveraging the benefits of proposed integral printing techniques, we investigated how the number of MagicTacs produced in a single batch affects both the printing cost and time. As displayed in Fig.~\ref{print_time_scene}(A), The total height of the MagicTac contact module is 25.5$mm$, and its inner height is 8.5$mm$, with a maximum diameter of 41$mm$. The print tray in the Stratasys J826 printer can be approximately divided into four areas (a1-a4), each covering a total size of 25$cm$ x 25$cm$. In each print area zone, the travel distance of the print nozzle remains relatively fixed, equivalent to the width of the print tray, regardless of the size of the printed part. 
This indicates that the time cost of producing a single and multiple MagicTacs within the same print area is similar. 
Also, it is closely linked to the maximum print height, a factor dictated by the principles of the additive printing technique.
The higher MagicTac designs necessitate longer printing times due to the increased number of layers required.
The results of average printing time and cost related to sensor production in a single printing batch are summarised in Fig.~\ref{print_time_scene} (C). 
Overall, the average printing time and cost for MagicTacs reduce with larger batch sizes, achieving minimums of 24.6 minutes and £4.76, respectively. However, the optimal printing time is realized at a batch size of 5, not 10, due to the printer's capacity to fit only up to 5 MagicTacs within a single print area. Exceeding this limit requires using additional print areas, which indirectly extends the print nozzles' travel distance, impacting time efficiency.

\begin{figure*}[!htbp]
	\centering
 \captionsetup{font=footnotesize,labelsep=period}
	\includegraphics[width = 1\hsize]{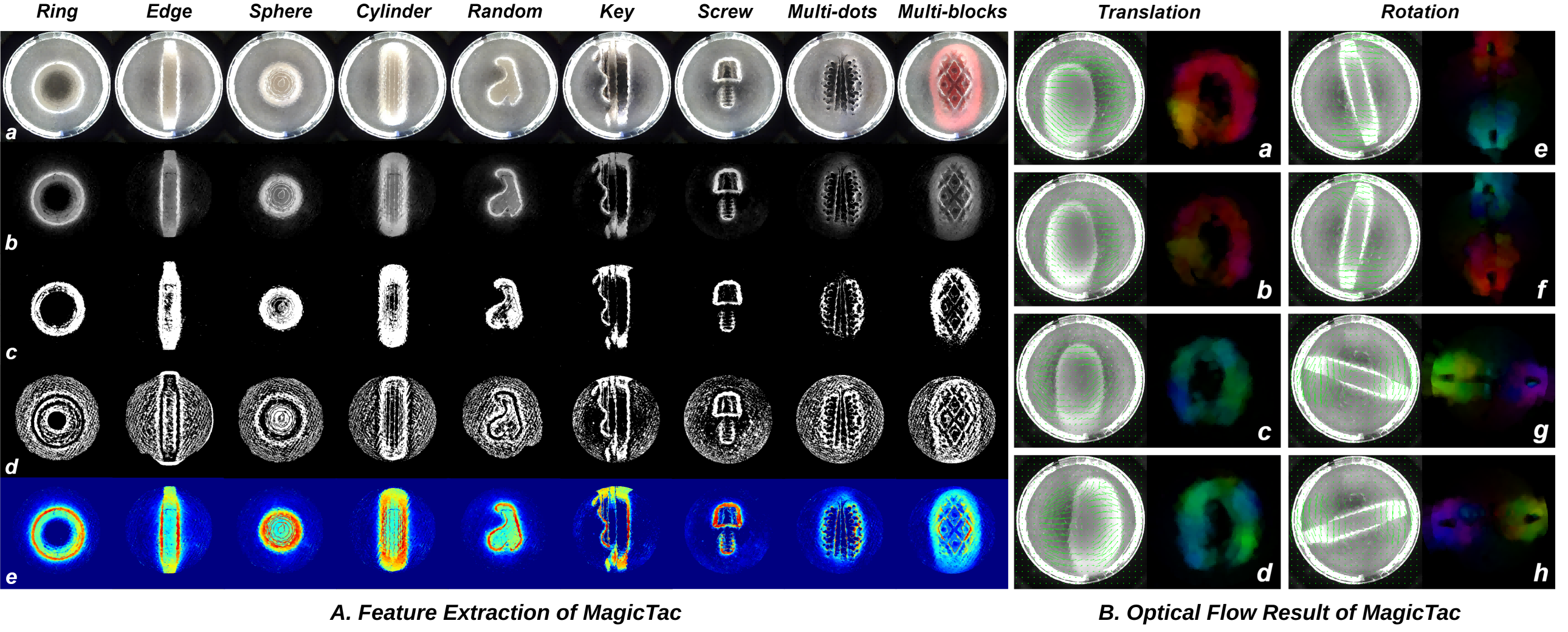}
	\caption{A: Tactile feature extractions on different objects, (a) raw image of MagicTac, (b) image difference, (c) contact region texture, (d) grid distribution mapping, (e) tactile change field. B: Optical flow analysis with translation and rotation situations, (a)-(d) optical flow changes due to horizontal translation in different directions and amplitudes, (e)-(h) optical flow changes caused by torsional movement in different directions and angles.}
 \vspace{-0.3cm}
	\label{optical_flow}
\end{figure*}

\section{Performance Evaluation}

\subsection{Tactile Reconstruction}

To evaluate the tactile sensing capabilities of MagicTac, five 3D-printed objects with distinct shapes and five real-world items with varying textures are used for experiments. Fig.~\ref{optical_flow} A.(a)-(b) showcases the raw images and their deviations from a baseline image. Upon contact with an object, the skin's edges illuminate more brightly, reflecting the internally projected light in regions undergoing notable deformation. This brightness increase at the edges serves as a visual indicator of the skin layer's interaction with the object, which represents areas of significant contact and deformation. 

Close to the contact point, the visibility of a grid-based projection pattern from the 3D multi-layer grid-based structure is enhanced, which facilitates the capture of comprehensive tactile information, including shear and torsion forces. This visibility is enabled by MagicTac's transparent property, which merges visual and tactile feedback while the object's color, grid-filled background, or other elements may affect the tactile data extraction. In this case, we decoupled the contact region texture and grid distribution mapping, as seen in Fig.~\ref{optical_flow} A.(c)-(d). This separation process allows for the identification of distinct features, including stepped print patterns on the surface of 3D-printed parts, diverse textures on the surfaces of real-world objects, and the elastic compression of the surrounding grids caused by the contacting objects. The final tactile change fields are shown in Fig.~\ref{optical_flow} A.(e), where the distribution of colors represents the deformation gradient in the contact area.


To further analyze the role of MagicTac design in dynamic contact tasks, we applied the optical flow method due to its capability to capture the deformations in grid structures from a global range. In Fig.~\ref{optical_flow} B.(a)-(d), if one presses the MagicTac skin to a certain depth with the thumb and moves it horizontally, the grid cells around the contact area will shift in the same direction due to their elasticity. This movement tendency is effectively captured by the optical flow method. The deeper the press, the larger the horizontal displacement, resulting in a more pronounced overall drift in the distribution of grids. Similar tests on rotation are illustrated in Fig.~\ref{optical_flow} B.(e)-(h), where the grid cells on either side of the twisted object rotate in the same direction.

\begin{figure*}[!htbp]
	\centering
 \captionsetup{font=footnotesize,labelsep=period}
	\includegraphics[width = 0.85\hsize]{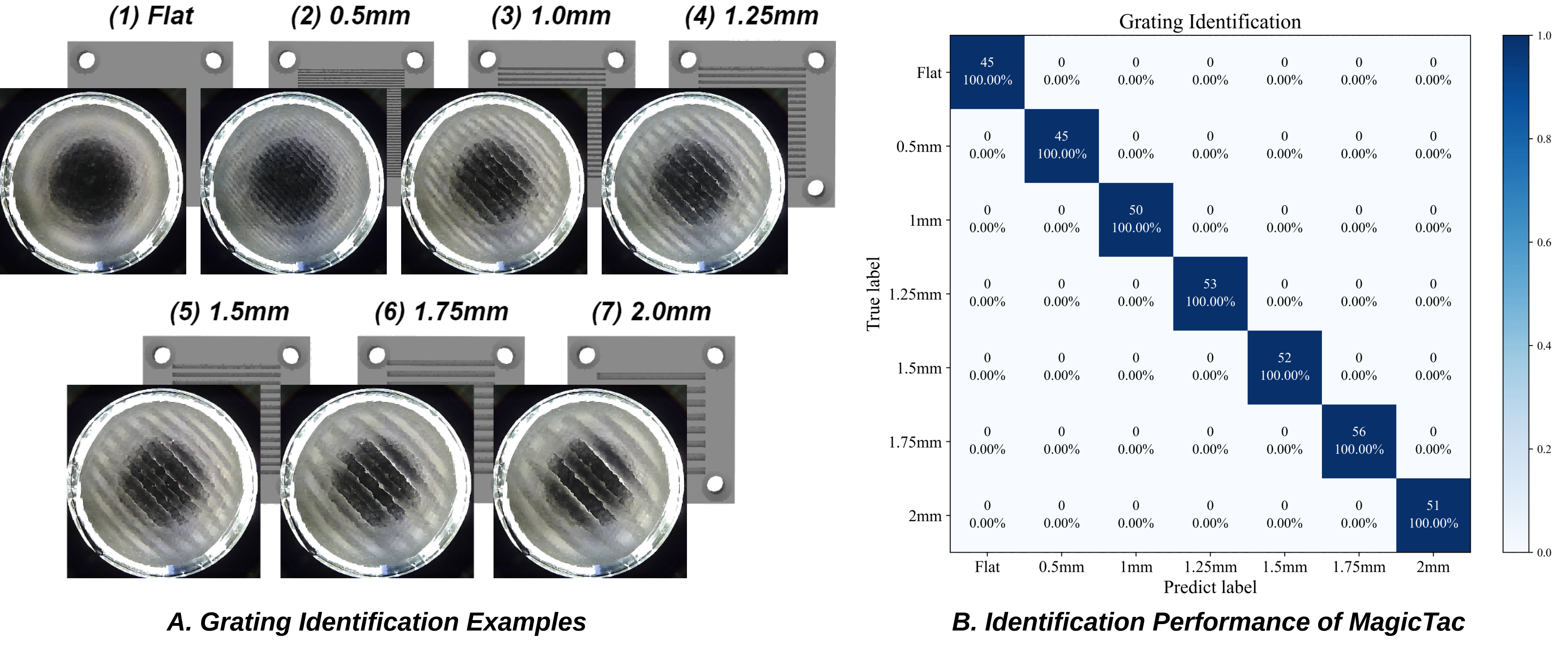}
	\caption{A: Grating boards used in grating identification task. B: Performance MagicTac on grating identification task.}
	\label{grating}
\end{figure*}

\begin{figure*}[!htbp]
	\centering
	\includegraphics[width = 0.85\hsize]{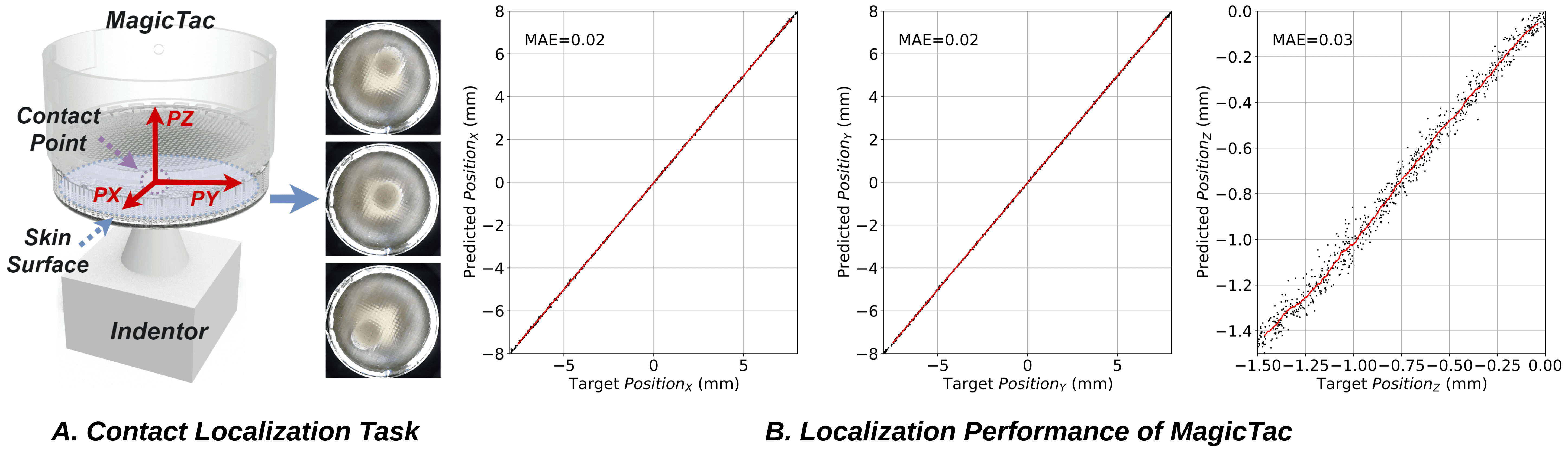}
 \captionsetup{font=footnotesize,labelsep=period}
	\caption{A: Experiment setup for contact localization task. B: Regression performance in contact localization task. }
 \vspace{-0.3cm}
	\label{localization}
\end{figure*}

\subsection{Grating Identification}

Grating identification can provide critical insights into the texture, quality, and even the type of material \cite{pestell2022artificial}. We applied a group of grating boards with different densities to evaluate the spatial resolution of the MagicTac. As displayed in Fig.~\ref{grating}(A), the grating boards are classified into 7 grades based on the grille spacing, including 0$mm$ (flat surface), 0.5$mm$, 1$mm$, 1.25$mm$, 1.5$mm$, 1.75$mm$, and 2$mm$. Since these grating boards have rigid flat surfaces, a customized curved-surface MagicTac was manufactured to ensure effective deformation between the sensor and board. The entire process only took less than 2 hours from design to ready for use, highlighting the advantage of integral printing. 




A desktop robot (Dobot MG400) was used to collect data in the method described in \cite{lepora2022digitac}, gathering 500 pieces of data for each grating board, for a total of 3,500 pieces. Among these data, 70\% were allocated to the training set, 20\% to the validation set, and the remaining 10\% constituted the test set. The classifier was trained end-to-end using a neural network model, specifically DenseNet121\cite{huang2017densely}. To realize the classification function, two additional fully connected layers were introduced to generate the final output. Results indicated that the MagicTac can handle all grating identification tasks, ranging from millimeter to quarter-millimeter, with a prediction accuracy of 100\%, as shown in Fig.~\ref{grating} (B). 


\subsection{Contact Localization}

Contact localization represents the capability of a sensor to precisely determine the point of contact when it interacts with an external object \cite{molchanov2016contact}. It requires the sensor to estimate the position of a single contact point, including plane coordinates along the $X$ and $Y$ axes, and the press depth along the $Z$ axis. This process can also be referred to as 3D position regression. As illustrated in Fig.~\ref{localization} (A), a 3D-printed cone with a tip diameter of 6mm is placed under the MagicTac as the indentor to generate a single-point contact. The contact position $X$/$Y$ can be accurately estimated by the plane center of the deformation area, whereas the prediction of depth $Z$ relies more on the distribution change of the embedded grid located around the deformation area. Similar to the grating identification task, we also use a desktop robot to collect data with the position [$PX$, $PY$, $PZ$] range of [[-8, -8, -1.5], [8, 8, 0]]($mm$). Considering the total thickness of the embedded grid within MagicTac can only reach 6$mm$, the pressing range in the $Z$-axis direction is limited to 1.5$mm$. Consequently, a total of 3000 images were collected. Of these 75\%  were allocated to the training set, and the remaining images were divided into the validation and test sets. The task of the contact localization regression continues to be undertaken by DenseNet121 \cite{huang2017densely}, whose output dimension changed to 3 for the 3D position estimation. The trained results of contact localization are summarised in Fig.~\ref{localization} (B). The MAE of position $X$ and position $Y$ are both 0.02$mm$. Meanwhile, the MAE of depth $Z$ is 0.03$mm$. Results indicate that MagicTac can accurately perceive and localize external contact.

\section{Conclusions and Future Work}
We propose MagicTac, a low-cost and easy-to-fabricate tactile sensor with a 3D multi-layer grid-based structure. Moreover, thanks to the integral printing technique (grid-based multi-material additive manufacturing technique), the fabrication of MagicTac not only eliminates the need for a manual fabrication process but also reduces its production cost to as low as £4.76, the manufacturing time to 24.6 minutes. Experimental results indicate that MagicTac can capture the intricate details of geometric shapes. Due to its unique embedded grid, MagicTac can readily detect both shear motions and rotations. The contact localization task demonstrates that MagicTac can accurately locate contact stimuli with an average horizontal localization error is 0.02$mm$, and an average indentation error is 0.03$mm$. In the future, we will integrate MagicTac closely with robotic hands to realize contact-rich dexterous manipulation.

\section*{Acknowledgment}
The authors would like to thank Andrew Stinchcombe, Ugnius Bajarunas, and Tom Barnes for their help in sensor design and fabrication.

\bibliographystyle{IEEEtran}
\bibliography{references}

\begin{thebibliography}{10}
\providecommand{\url}[1]{#1}
\csname url@samestyle\endcsname
\providecommand{\newblock}{\relax}
\providecommand{\bibinfo}[2]{#2}
\providecommand{\BIBentrySTDinterwordspacing}{\spaceskip=0pt\relax}
\providecommand{\BIBentryALTinterwordstretchfactor}{4}
\providecommand{\BIBentryALTinterwordspacing}{\spaceskip=\fontdimen2\font plus
\BIBentryALTinterwordstretchfactor\fontdimen3\font minus \fontdimen4\font\relax}
\providecommand{\BIBforeignlanguage}[2]{{%
\expandafter\ifx\csname l@#1\endcsname\relax
\typeout{** WARNING: IEEEtran.bst: No hyphenation pattern has been}%
\typeout{** loaded for the language `#1'. Using the pattern for}%
\typeout{** the default language instead.}%
\else
\language=\csname l@#1\endcsname
\fi
#2}}
\providecommand{\BIBdecl}{\relax}
\BIBdecl

\bibitem{li2023marker}
M.~Li, T.~Li, and Y.~Jiang, ``Marker displacement method used in vision-based tactile sensors—from 2d to 3d-a review,'' \emph{IEEE Sensors Journal}, 2023.

\bibitem{scharff2022rapid}
R.~B. Scharff, D.-J. Boonstra, L.~Willemet, X.~Lin, and M.~Wiertlewski, ``Rapid manufacturing of color-based hemispherical soft tactile fingertips,'' in \emph{2022 IEEE 5th International Conference on Soft Robotics (RoboSoft)}.\hskip 1em plus 0.5em minus 0.4em\relax IEEE, 2022, pp. 896--902.

\bibitem{lepora2021soft}
N.~F. Lepora, ``Soft biomimetic optical tactile sensing with the tactip: A review,'' \emph{IEEE Sensors Journal}, vol.~21, no.~19, pp. 21\,131--21\,143, 2021.

\bibitem{yuan2017gelsight}
W.~Yuan, S.~Dong, and E.~H. Adelson, ``Gelsight: High-resolution robot tactile sensors for estimating geometry and force,'' \emph{Sensors}, vol.~17, no.~12, p. 2762, 2017.

\bibitem{gomes2020geltip}
D.~F. Gomes, Z.~Lin, and S.~Luo, ``Geltip: A finger-shaped optical tactile sensor for robotic manipulation,'' in \emph{2020 IEEE/RSJ International Conference on Intelligent Robots and Systems (IROS)}.\hskip 1em plus 0.5em minus 0.4em\relax IEEE, 2020, pp. 9903--9909.

\bibitem{taylor2022gelslim}
I.~H. Taylor, S.~Dong, and A.~Rodriguez, ``Gelslim 3.0: High-resolution measurement of shape, force and slip in a compact tactile-sensing finger,'' in \emph{2022 International Conference on Robotics and Automation (ICRA)}.\hskip 1em plus 0.5em minus 0.4em\relax IEEE, 2022, pp. 10\,781--10\,787.

\bibitem{sato2009finger}
K.~Sato, K.~Kamiyama, N.~Kawakami, and S.~Tachi, ``Finger-shaped gelforce: sensor for measuring surface traction fields for robotic hand,'' \emph{IEEE Transactions on Haptics}, vol.~3, no.~1, pp. 37--47, 2009.

\bibitem{lin2023dtact}
C.~Lin, Z.~Lin, S.~Wang, and H.~Xu, ``Dtact: A vision-based tactile sensor that measures high-resolution 3d geometry directly from darkness,'' in \emph{2023 IEEE International Conference on Robotics and Automation (ICRA)}.\hskip 1em plus 0.5em minus 0.4em\relax IEEE, 2023, pp. 10\,359--10\,366.

\bibitem{lambeta2020digit}
M.~Lambeta, P.-W. Chou, S.~Tian, B.~Yang, B.~Maloon, V.~R. Most, D.~Stroud, R.~Santos, A.~Byagowi, G.~Kammerer \emph{et~al.}, ``Digit: A novel design for a low-cost compact high-resolution tactile sensor with application to in-hand manipulation,'' \emph{IEEE Robotics and Automation Letters}, vol.~5, no.~3, pp. 3838--3845, 2020.

\bibitem{padmanabha2020omnitact}
A.~Padmanabha, F.~Ebert, S.~Tian, R.~Calandra, C.~Finn, and S.~Levine, ``Omnitact: A multi-directional high-resolution touch sensor,'' in \emph{2020 IEEE International Conference on Robotics and Automation (ICRA)}.\hskip 1em plus 0.5em minus 0.4em\relax IEEE, 2020, pp. 618--624.

\bibitem{romero2020soft}
B.~Romero, F.~Veiga, and E.~Adelson, ``Soft, round, high resolution tactile fingertip sensors for dexterous robotic manipulation,'' in \emph{2020 IEEE International Conference on Robotics and Automation (ICRA)}.\hskip 1em plus 0.5em minus 0.4em\relax IEEE, 2020, pp. 4796--4802.

\bibitem{do2023densetact}
W.~K. Do, B.~Jurewicz, and M.~Kennedy, ``Densetact 2.0: Optical tactile sensor for shape and force reconstruction,'' in \emph{2023 IEEE International Conference on Robotics and Automation (ICRA)}.\hskip 1em plus 0.5em minus 0.4em\relax IEEE, 2023, pp. 12\,549--12\,555.

\bibitem{zhang2023gelstereo}
C.~Zhang, S.~Cui, S.~Wang, J.~Hu, Y.~Cai, R.~Wang, and Y.~Wang, ``Gelstereo 2.0: An improved gelstereo sensor with multimedium refractive stereo calibration,'' \emph{IEEE Transactions on Industrial Electronics}, 2023.

\bibitem{yang2021enhanced}
Y.~Yang, X.~Wang, Z.~Zhou, J.~Zeng, and H.~Liu, ``An enhanced fingervision for contact spatial surface sensing,'' \emph{IEEE Sensors Journal}, vol.~21, no.~15, pp. 16\,492--16\,502, 2021.

\bibitem{lepora2022digitac}
N.~F. Lepora, Y.~Lin, B.~Money-Coomes, and J.~Lloyd, ``Digitac: A digit-tactip hybrid tactile sensor for comparing low-cost high-resolution robot touch,'' \emph{IEEE Robotics and Automation Letters}, vol.~7, no.~4, pp. 9382--9388, 2022.

\bibitem{lin2019sensing}
X.~Lin and M.~Wiertlewski, ``Sensing the frictional state of a robotic skin via subtractive color mixing,'' \emph{IEEE Robotics and Automation Letters}, vol.~4, no.~3, pp. 2386--2392, 2019.

\bibitem{lin2020curvature}
X.~Lin, L.~Willemet, A.~Bailleul, and M.~Wiertlewski, ``Curvature sensing with a spherical tactile sensor using the color-interference of a marker array,'' in \emph{2020 IEEE International Conference on Robotics and Automation (ICRA)}.\hskip 1em plus 0.5em minus 0.4em\relax IEEE, 2020, pp. 603--609.

\bibitem{yamaguchi2016combining}
A.~Yamaguchi and C.~G. Atkeson, ``Combining finger vision and optical tactile sensing: Reducing and handling errors while cutting vegetables,'' in \emph{2016 IEEE-RAS 16th International Conference on Humanoid Robots (Humanoids)}.\hskip 1em plus 0.5em minus 0.4em\relax IEEE, 2016, pp. 1045--1051.

\bibitem{yamaguchi2021fingervision}
A.~Yamaguchi, ``Fingervision with whiskers: Light touch detection with vision-based tactile sensors,'' in \emph{2021 Fifth IEEE International Conference on Robotic Computing (IRC)}.\hskip 1em plus 0.5em minus 0.4em\relax IEEE, 2021, pp. 56--64.

\bibitem{abad2020visuotactile}
A.~C. Abad and A.~Ranasinghe, ``Visuotactile sensors with emphasis on gelsight sensor: A review,'' \emph{IEEE Sensors Journal}, vol.~20, no.~14, pp. 7628--7638, 2020.

\bibitem{zhang2022hardware}
S.~Zhang, Z.~Chen, Y.~Gao, W.~Wan, J.~Shan, H.~Xue, F.~Sun, Y.~Yang, and B.~Fang, ``Hardware technology of vision-based tactile sensor: A review,'' \emph{IEEE Sensors Journal}, 2022.

\bibitem{lin20239dtact}
C.~Lin, H.~Zhang, J.~Xu, L.~Wu, and H.~Xu, ``9dtact: A compact vision-based tactile sensor for accurate 3d shape reconstruction and generalizable 6d force estimation,'' \emph{arXiv preprint arXiv:2308.14277}, 2023.

\bibitem{wang2021gelsight}
S.~Wang, Y.~She, B.~Romero, and E.~Adelson, ``Gelsight wedge: Measuring high-resolution 3d contact geometry with a compact robot finger,'' in \emph{2021 IEEE International Conference on Robotics and Automation (ICRA)}.\hskip 1em plus 0.5em minus 0.4em\relax IEEE, 2021, pp. 6468--6475.

\bibitem{jiang20183}
H.~Jiang, Y.~Yan, X.~Zhu, and C.~Zhang, ``A 3-d surface reconstruction with shadow processing for optical tactile sensors,'' \emph{Sensors}, vol.~18, no.~9, p. 2785, 2018.

\bibitem{winstone2013tactip}
B.~Winstone, G.~Griffiths, T.~Pipe, C.~Melhuish, and J.~Rossiter, ``Tactip-tactile fingertip device, texture analysis through optical tracking of skin features,'' in \emph{Biomimetic and Biohybrid Systems: Second International Conference, Living Machines 2013, London, UK, July 29--August 2, 2013. Proceedings 2}.\hskip 1em plus 0.5em minus 0.4em\relax Springer, 2013, pp. 323--334.

\bibitem{sakuma2018universal}
T.~Sakuma, F.~Von~Drigalski, M.~Ding, J.~Takamatsu, and T.~Ogasawara, ``A universal gripper using optical sensing to acquire tactile information and membrane deformation,'' in \emph{2018 IEEE/RSJ International Conference on Intelligent Robots and Systems (IROS)}.\hskip 1em plus 0.5em minus 0.4em\relax IEEE, 2018, pp. 1--9.

\bibitem{ward2017exploiting}
B.~Ward-Cherrier, L.~Cramphorn, and N.~F. Lepora, ``Exploiting sensor symmetry for generalized tactile perception in biomimetic touch,'' \emph{IEEE Robotics and Automation Letters}, vol.~2, no.~2, pp. 1218--1225, 2017.

\bibitem{zhang2023novel}
S.~Zhang, Y.~Yang, J.~Shan, F.~Sun, and B.~Fang, ``A novel vision-based tactile sensor using lamination and gilding process for improvement of outdoor detection and maintainability,'' \emph{IEEE Sensors Journal}, vol.~23, no.~4, pp. 3558--3566, 2023.

\bibitem{pestell2022artificial}
N.~Pestell, T.~Griffith, and N.~F. Lepora, ``Artificial sa-i and ra-i afferents for tactile sensing of ridges and gratings,'' \emph{Journal of the Royal Society Interface}, vol.~19, no. 189, p. 20210822, 2022.

\bibitem{huang2017densely}
G.~Huang, Z.~Liu, L.~Van Der~Maaten, and K.~Q. Weinberger, ``Densely connected convolutional networks,'' in \emph{Proceedings of the IEEE conference on computer vision and pattern recognition}, 2017, pp. 4700--4708.

\bibitem{molchanov2016contact}
A.~Molchanov, O.~Kroemer, Z.~Su, and G.~S. Sukhatme, ``Contact localization on grasped objects using tactile sensing,'' in \emph{2016 IEEE/RSJ International Conference on Intelligent Robots and Systems (IROS)}.\hskip 1em plus 0.5em minus 0.4em\relax IEEE, 2016, pp. 216--222.

\end{thebibliography}

\end{document}